\documentclass[12pt, letter]{article}

\usepackage[utf8]{inputenc}%
\usepackage[margin=1.2in,letterpaper,centering]{geometry}%
\usepackage{circuitikz}
\usepackage{subcaption}
\usepackage{todonotes}
\usepackage{graphicx}%
\usepackage{multirow}%
\usepackage{amsmath,amssymb,amsfonts}%
\usepackage{amsthm}%
\usepackage{mathrsfs}%
\usepackage[title]{appendix}%
\usepackage{xcolor}%
\usepackage{textcomp}%
\usepackage{manyfoot}%
\usepackage{booktabs}%
\usepackage{algorithm}%
\usepackage{algorithmicx}%
\usepackage{algpseudocode}%
\usepackage{listings}%
\usepackage{pdflscape}%
\usepackage{adjustbox}%
\usepackage{cite}%

\usepackage{hyperref}%

\begin{document}
\begin{center}
{\LARGE\bfseries 
Symbolic Regression with\\[4pt]
Multimodal Large Language Models\\[10pt]
and Kolmogorov–Arnold Networks}
\end{center}
\begin{center}
Thomas R.\ Harvey$^{a,}$\footnote{\href{mailto:trharvey@mit.edu}{trharvey@mit.edu}}, 
Fabian Ruehle$^{b,c,a,}$\footnote{\href{mailto:f.ruehle@northeastern.edu}{f.ruehle@northeastern.edu}}, 
Kit Fraser-Taliente$^{d,}$\footnote{\href{mailto:cristofero.fraser-taliente@physics.ox.ac.uk}{cristofero.fraser-taliente@physics.ox.ac.uk}}, \\
James Halverson$^{b,a,}$\footnote{\href{mailto:j.halverson@northeastern.edu}{j.halverson@northeastern.edu}} \\[10mm]

\bigskip
{
	{\it ${}^{\text{a}}$ NSF AI Institute for Fundamental Interactions, MIT, Cambridge, MA 02139, USA}\\[.5em]
	{\it ${}^{\text{b}}$ Department of Physics, Northeastern University, Boston, MA 02115, USA}\\[.5em]
	{\it ${}^{\text{c}}$ Department of Mathematics, Northeastern University, Boston, MA 02115, USA}\\[.5em]
	{\it ${}^{\text{d}}$ Rudolf Peierls Centre for Theoretical Physics, University of Oxford, Oxford OX1 2JD, UK}\\[.5em]
}
\end{center}
\setcounter{footnote}{0} 
\bigskip\bigskip

\begin{abstract}
We present a novel approach to symbolic regression using vision-capable large language models (LLMs) and the ideas behind Google DeepMind's \texttt{Funsearch}. The LLM is given a plot of a univariate function and tasked with proposing an ansatz for that function. The free parameters of the ansatz are fitted using standard numerical optimisers, and a collection of such ansätze make up the population of a genetic algorithm. Unlike other symbolic regression techniques, our method does not require the specification of a set of functions to be used in regression, but with appropriate prompt engineering, we can arbitrarily condition the generative step. By using Kolmogorov–Arnold Networks (KANs), we demonstrate that ``univariate is all you need'' for symbolic regression, and extend this method to multivariate functions by learning the univariate function on each edge of a trained KAN. The combined expression is then simplified by further processing with a language model.
\end{abstract}

%\keywords{Machine Learning~(ML), Symbolic Regression, Large Language Models~(LLM), Kolmogorov-Arnold Networks~(KANs), Interpretability, Funsearch}

\newpage
\tableofcontents

\section{Introduction}\label{sec:Intro}
Symbolic regression is a long-standing and inherently challenging problem in the fields of machine learning and applied mathematics. The task involves searching for mathematical expressions that best describe a given dataset, yet the space of possible functions is unyielding. To navigate this expansive search space effectively, modern symbolic regression algorithms often rely on explicit or implicit complexity measures to mitigate expression ``bloat'' and encourage simpler expressions~\cite{pysr,operon,Bartlett:2022kyi}.

Humans exhibit an impressive intuitive ability to infer plausible functional forms from visual representations of data, typically guided by a natural simplicity bias. When presented with a graph of a function, expert human observers tend to prefer and propose simpler ansätze that often outperform those generated by automated symbolic regression methods.

This observation naturally raises the question: can large multimodal language models (LLMs) emulate this human-like intuition? In this work, we show that they can---at least in the context of univariate functions---especially when integrated with DeepMind's \texttt{FunSearch}~\cite{romera2024mathematical}. To that end, we introduce LLM-LEx (Large Language Models Learning Expressions), a package available at our GitHub repository~\cite{llmsr}, which leverages commercially available LLMs for guiding symbolic regression.

By invoking the Kolmogorov–Arnold representation theorem, which tells us that multivariate functions can be represented as sums and compositions of univariate functions, we moreover argue that focussing on univariate cases is sufficient for a broad class of problems. We train Kolmogorov–Arnold Networks (KANs) on the data for which we aim to find a symbolic expression. Then, we employ LLM-LEx to propose symbolic expressions for the edge functions within these networks. The combined expression is then simplified using an additional step of generative modelling using LLMs. We refer to this combined method as KAN-LEx, and it is likewise publicly available at our repository.

We emphasise that we do not expect beyond state-of-the-art performance from this method of symbolic regression. For one, many traditional and freely available packages have been highly optimised through years of development. Our aim is simply to demonstrate that this new approach to symbolic regression is viable and surprisingly successful given the simplicity of its implementation. Our initial implementation comprised approximately 100 lines of code. The combination of our method with KANs allows application to multivariate functions, although in principle any symbolic regression method can be used for this step.

Unless stated otherwise, we use \texttt{gpt-4o} via OpenRouter throuhout.
\section{Symbolic Regression with LLMs}\label{sec:LLMLEX}
In this section, we begin with a simple example of using a vision-capable language model for symbolic regression with a one-shot prompt. We then extend this approach by incorporating a genetic algorithm. We compare to traditional techniques and subsequently benchmark against a set of randomly generated functions. We also consider the use of open-source language models as an alternative to proprietary ones. We then consider the effects of noisy data, before concluding with a discussion on prompt engineering and learning special functions. 
\subsection{A simple example}
\begin{figure}[t]
    \centering
    \includegraphics[width=0.8\linewidth]{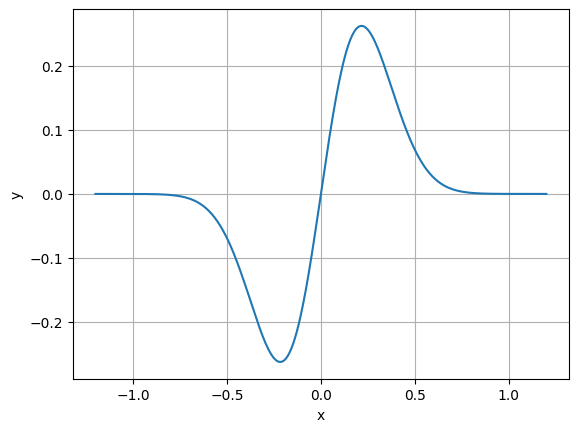}
    \caption{An example function to learn}
    \label{fig:SimpleExample}
\end{figure} 
Consider the graph in Figure~\ref{fig:SimpleExample}, generated by interpolating 500 evenly spaced points between $-1.2$ and $1.2$. It is not too difficult for a human to make a reasonable ansatz for the underlying function. In particular, the oscillatory behaviour and intersection at zero suggest the function may take the form $y(x) = g(x)\sin(a x)$, where $g(x)$ is a smooth function and $a$ is a constant. The way the function flattens out toward zero then suggests the presence of exponential decay. Putting these together, a suitable, and correct, ansatz for such a function would be
\begin{equation}
    y(x) = c e^{- b x^2} \sin(a x),
\end{equation}
which after fitting to the data yields $a=2$, $b=10$ and $c=1$. Despite the apparent ease with which a human can make an educated guess for the function, developing software that replicates this kind of intuition is highly challenging. This, in essence, captures the core difficulty of symbolic regression. For example, when we input the 100 data points into a traditional symbolic regression algorithm---specifically, Mathematica's {\texttt{FindFormula}}\footnote{As it is proprietary, very little public information is available about how {\texttt{FindFormula}} operates. However, it is believed to involve a heuristic search over symbolic expressions, guided by both the complexity and accuracy of the functions. Such methodology is typical of other approaches to symbolic regression.} function~\cite{Mathematica}---we find
\begin{equation}
\label{eq:mathmfit}
\begin{aligned}
    f(x) = &-13.7072 x^{19.}+104.411 x^{17.}-344.585 x^{15.}+647.044 x^{13.}-763.814 x^{11.}\\ &+ 591.433 x^{9.}-303.867 x^{7.}+101.756 x^{5.}-20.6594 x^{3.}+1.98844 x.
 \end{aligned}
\end{equation}
While \eqref{eq:mathmfit} gives a good fit to the data, for the given range, the resulting expression is difficult to interpret and offers little additional insight into the underlying structure of the problem: we might just as well have fitted the data with a high-degree polynomial from the outset.

We turn instead to a multimodal language model (in this example, \texttt{gpt-4o}): we will see language models display some of the intuitive symbolic regression ability exhibited by humans, most likely acquired from their extensive pretraining data. We present the LLM with the image in Figure~\ref{fig:SimpleExample}, accompanied by the following prompt:
\begin{center}
\begin{it}
    ``An initial ansatz for this function is 
    % \begin{lstlisting}[language=Python]
    curve = lambda x, params: params[0]*x + params[1].
% \end{lstlisting}
    Give an improved ansatz for the image. params can be any length''
\end{it}  
\end{center}

The response from the LLM will, sometimes\footnote{As the response from the LLM is probabilistic, it does not always return exactly the correct function. We address this in the next subsection by incorporating a genetic algorithm, where this actually becomes a strength. The probabilistic nature of the LLM's response can be thought of as mutation; when given access to LLM sampling parameters, we gain some parametric control over this mutation.}, contain the Python lambda function:

\begin{lstlisting}[language=Python]
    curve = lambda x, params:
    np.sin(params[0]*x)*np.exp(-params[1]*x**2)
\end{lstlisting}
Remarkably, this is exactly the functional form we were aiming for. We propose adopting this as the foundation of a new methodology for symbolic regression. As the range is increased, and the exponential decay becomes clearer, the LLM predicts the correct function more frequently, whilst Mathematica resorts to suggesting the function is zero.

\subsection{Extending to a genetic algorithm and \texttt{Funsearch}}
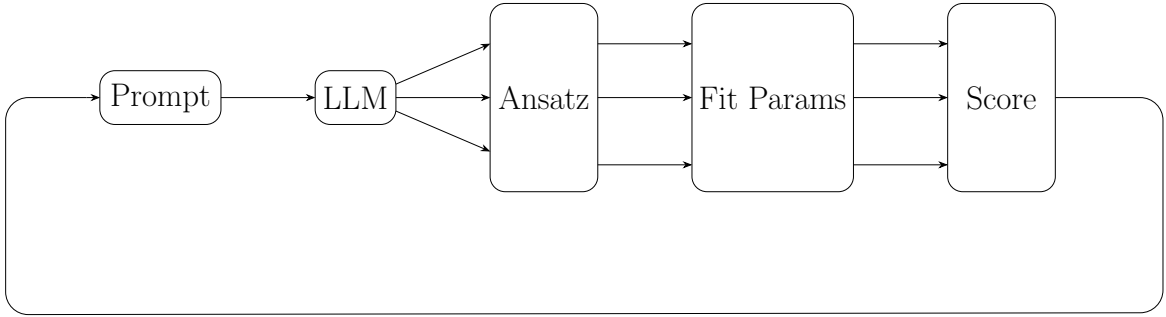
\begin{figure}[!ht]
\centering
\resizebox{1\textwidth}{!}{%
\begin{circuitikz}
\tikzstyle{every node}=[font=\LARGE]

% Labels
\node [font=\Large] at (5,8.25) {LLM};

% Prompt box
\draw[rounded corners=8pt] (0.25,7.75) rectangle node {\Large Prompt} (2.5,8.75);

% Pre-LLM box
\draw[rounded corners=8pt] (4.25,7.75) rectangle (5.75,8.75);

% LLM arrows
\draw[->, >=Stealth] (5.75,8.25) -- (7.5,8.25);  % To Ansatz (middle)
\draw[->, >=Stealth] (5.75,8.5) -- (7.5,9.25);   % To Ansatz (top)
\draw[->, >=Stealth] (5.75,8) -- (7.5,7.25);     % To Ansatz (bottom)

% Prompt to LLM
\draw[->, >=Stealth] (2.5,8.25) -- (4.25,8.25);

% Ansatz box
\draw[rounded corners=8pt] (7.5,6.5) rectangle node {\Large Ansatz} (9.5,10);

% Ansatz to Fit Params
\draw[->, >=Stealth] (9.5,9.25) -- (11.25,9.25);  % Top
\draw[->, >=Stealth] (9.5,8.25) -- (11.25,8.25);  % Mid
\draw[->, >=Stealth] (9.5,7) -- (11.25,7);        % Bottom

% Fit Params box
\draw[rounded corners=8pt] (11.25,6.5) rectangle node {\Large Fit Params} (14.25,10);

% Fit Params to Score
\draw[->, >=Stealth] (14.25,9.25) -- (16,9.25);  % Top
\draw[->, >=Stealth] (14.25,8.25) -- (16,8.25);  % Mid
\draw[->, >=Stealth] (14.25,7) -- (16,7);        % Bottom

% Score box
\draw[rounded corners=8pt] (16,6.5) rectangle node {\Large Score} (18,10);

% Fully rounded feedback loop
\draw[rounded corners=12pt]
  (18,8.25) -- (20,8.25)
  -- (20,4.25)
  -- (-1.5,4.215)
  -- (-1.5,8.25)
  -- (0,8.25);

% Final arrow into Prompt
\draw[->, >=Stealth] (0,8.25) -- (0.25,8.25);

\end{circuitikz}
}%
\caption{The structure of LLM-LEx.}
\label{fig:llmlex}
\end{figure}
Despite the success demonstrated in the example above, language models struggle to produce satisfactory results when presented with more complex functions. To address this limitation, we propose enhancing the methodology by incorporating a genetic algorithm~\cite{cantu1998survey}. The idea of using a genetic algorithm to generate prompts for a language model has been previously explored in the context of \texttt{Funsearch}~\cite{romera2024mathematical,vonHippel:2025okr,ellenberg2025generativemodelingmathematicaldiscovery}. We suggest integrating this approach into the framework introduced in the previous section. The general structure of our approach is indicated in Fig.~\ref{fig:llmlex}.

We begin with a population of proposed functions, all initially defined as the constant function.
\begin{lstlisting}[language=Python]
    lambda x,*params: params[0].
\end{lstlisting}
Each function in the population is evaluated using the following scoring metric:
\begin{equation}
    {\textbf{Score}}(f_\theta) =\frac{1}{N} \sum_{i=1}^N\frac{|f_\theta(x_i) - y_i|^2}{\text{max}(\alpha|y_i|,\text{global-scale})^2},
\end{equation}
where $(x_i, y_i)$ are the data points, $\alpha$ is a hyperparameter (default value: 0.01), and $f$ is the candidate function. The parameters $\theta$ are optimised using \texttt{SciPy}'s optimisation routines. The value of global-scale is a non-vanishing characteristic scale for the function. It is defined as:
\begin{align}
\begin{split}
    \text{global-scale} &= \text{max}(\{\text{MAD}_y,\alpha\cdot\text{mean}(\{|y_i|\},\epsilon)\}),\\
    \text{MAD}_y &= \text{median}(\{y_i - \text{median}(\{y_j\})\})
\end{split}
\end{align}
where $\epsilon \ll 1$ is a small hyperparameter which we introduce to avoid division by zero. The intuition behind this scoring metric is that, unlike mean squared error (MSE) (which favours fitting regions with large values) this score emphasises capturing the overall {\it shape} of the function.

The resulting scores are then normalised so that the maximum score in the population is unity; these normalised scores are denoted as $\{s_i \mid i = 1, \ldots, N\}$.

To construct the prompts for generating the next population of functions, two examples are randomly selected from the previous generation at least $N$ times (some may fail to parse, and so are redrawn). The selection is based on a probability distribution derived from the normalised scores, which are passed through a softmax function with temperature $T$. Specifically, the probability of choosing the $i$-th function is given by: 
\begin{equation}
    P_i = \frac{e^{s_i/T}}{\sum_j e^{s_j/T}},
\end{equation}
where, unless stated otherwise, $T=1$.

The two chosen functions are then used to construct the user prompt:
\begin{lstlisting}[language=Python]
    import numpy as np 
    curve_0 = lambda x,*params:<First Random Function>
    curve_1 = lambda x,*params:<Second Random Function>
    curve_2 = lambda x,*params:
\end{lstlisting}
along with the image to which we wish to fit. The system prompt is set as follows:
\begin{it}
\begin{center}
    ``You are a symbolic regression expert. Analyze the data in the image and provide an improved mathematical ansatz.
     Respond with ONLY the ansatz formula, without any explanation or commentary. Ensure it is in valid Python. You may use numpy functions. 
     params is a list of parameters that can be of any length or complexity.''
\end{center}
\end{it}
The function for the new population can then be extracted from the LLM's response. If the response fails to parse correctly, a new pair of functions is selected from the previous population. The genetic algorithm terminates early if any individual in the population exceeds a specified score threshold (by default, $10^{-5}$).

Whilst not discussed here, standard genetic algorithm techniques, such as `island models' and `elitism', can be straightforwardly incorporated into the process described above. This algorithm is implemented in a GitHub repository~\cite{llmsr}. From this point forward, we will refer to this algorithm as LLM-LEx (Large Language Models Learning Expressions).

\subsection{Comparison to traditional methods and benchmarking}
Many traditional methods for symbolic regression exist, typically involving the breeding and mutation of expression trees via genetic algorithms~\cite{pysr, operon, review_sym_reg, gp_book}. The specificities--- functions considered, as well as the techniques used to suppress the dominance of overly complex functions (a problem often referred to as ``bloat'')---vary depending on the implementation. A notable exception to this is ``exhaustive symbolic regression,'' where an exhaustive search is conducted over expressions up to a predefined maximal complexity~\cite{Bartlett:2022kyi}. As an example, we compare Mathematica's \texttt{FindFormula} function to our method~\cite{Mathematica}.

Whilst the use of LLMs for symbolic regression is not entirely new~\cite{shojaee2024llm,merler2024context,grayeli2024symbolic,li2024visymre}, existing methods either rely solely on raw data or use the LLM as an assistant (with both data and visual input) within a more traditional symbolic regression process. In contrast, our method exclusively uses the image as input, without any raw data. The model then infers relationships and patterns purely from visual information.

One key distinction between LLM-LEx (along with some of the other approaches using language models) and traditional methods is that LLM-LEx does not require the user to specify a list of basis functions. Instead, the language model selects appropriate functions based on those it has encountered during training. This approach mirrors how humans perform symbolic regression, with the added advantage that it is particularly easy to {\it condition} the generative steps by simply providing additional context to the prompted model (in, for example, the system prompt).

\begin{figure}[t]
    \centering
    \begin{subfigure}[b]{0.45\textwidth}
        \centering
        \includegraphics[width=\textwidth]{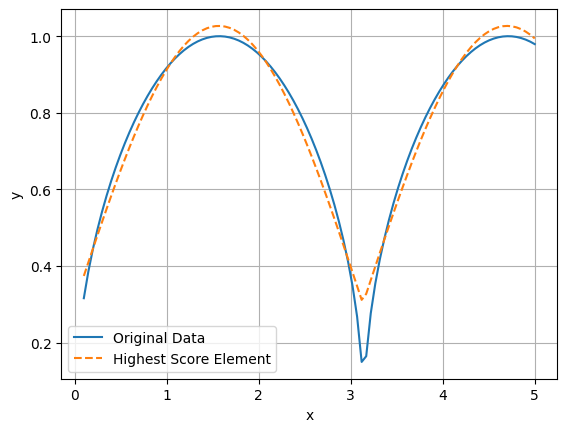}
        \caption{Target Function: $\sqrt{| \sin (x)| }$}
        \label{fig:fig1}
    \end{subfigure}
    \hfill
    \begin{subfigure}[b]{0.45\textwidth}
        \centering
        \includegraphics[width=\textwidth]{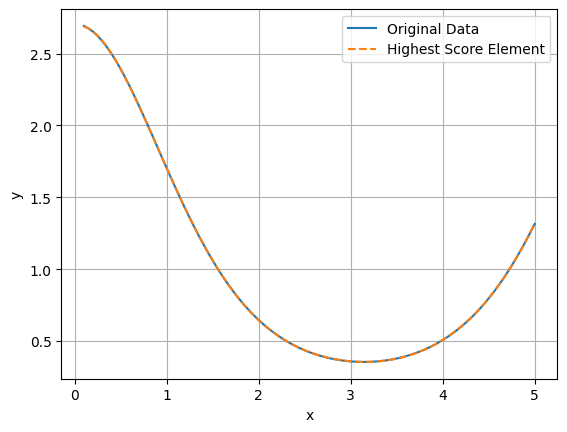}
        \caption{Target Function: $e^{\cos (x)}-0.0126997$ }
        \label{fig:fig2}
    \end{subfigure}
    \vskip\baselineskip
    \begin{subfigure}[b]{0.45\textwidth}
        \centering
        \includegraphics[width=\textwidth]{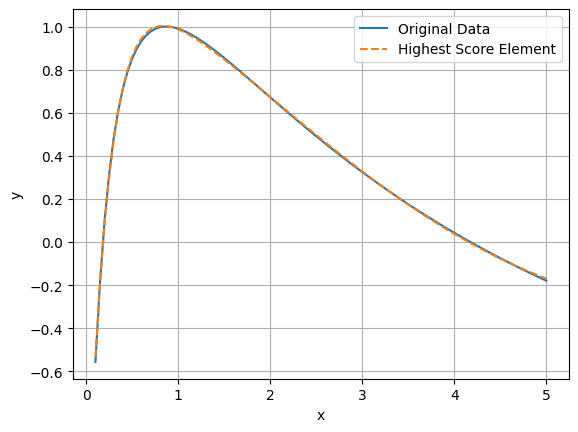}
        \caption{Target Function: $\sin \left(\log \left(\frac{4.1746}{x}\right)\right)$}
        \label{fig:fig3}
    \end{subfigure}
    \hfill
    \begin{subfigure}[b]{0.45\textwidth}
        \centering
        \includegraphics[width=\textwidth]{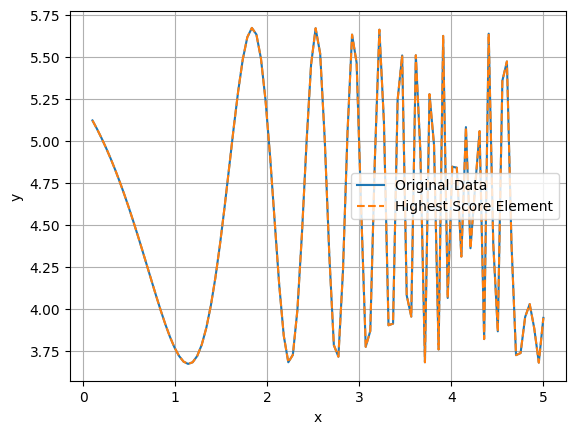}
        \caption{Target Function: $\cos \left(e^x\right)+4.67315$}
        \label{fig:fig4}
    \end{subfigure}
    \caption{Four example functions found by LLM-LEx with various scores. The functions returned by the algorithm and their scores are given in Table~\ref{tab:Compare}  and Table~\ref{tab:CompareScores}, respectively. The number of points used to produce the plots are the same that we used to provide the graph of the function to LLM-LEx, which leads to the aliasing effect in the highly oscillatory part of the last function.}
    \label{fig:examples}
\end{figure}

A comparison of the two methods can be found in Tables~\ref{tab:Compare} and~\ref{tab:CompareScores} for a set of randomly generated functions. Not only does LLM-LEx find the exact expression more frequently than Mathematica, but in all but one instance where Mathematica achieves a higher score, it returned a polynomial, which provides little additional insight into the nature of the function. Plots of some of the functions found by LLM-LEx are shown in Fig.~\ref{fig:examples}. Interestingly, even when the functions score poorly, they often appear aesthetically correct when plotted. The algorithm also seemed unaffected by the aliasing visible in the function $\cos \left(e^x\right) + 4.67315$. The functions used for benchmarking were generated from random symbolic trees in Mathematica. The code to generate these functions is available on the GitHub repository, in the file named \texttt{generate\_functions.nb}.

The primary downside of LLM-LEx over traditional methods is the cost of inference, and the latency of LLM calls. We hope that both of these factors will improve as the technology evolves. Currently, a population of 25 individuals over 10 generations takes about 10 minutes (although this can be muched improved using asynchronous API calls, a feature of LLM-LEx) and costs approximately \$0.50 when using the latest version of \texttt{gpt-4o}. The algorithm can terminate early if it reaches the required exit condition on its score; we naturally observe this in some cases. Additionally, the method can be made more cost-effective, typically at the expensive of some quality, by running one of the many available open-source models locally. We explore this option in the next section.

\begin{landscape}
\begin{table}[t]
    % \centering
\begin{adjustbox}{width=\columnwidth,center}
\begin{tabular}{|c|c|c|c|}
\hline
{\bf Expression} & {\bf Mathematica Result}  & {\bf LLM-LEx Result} & {\bf \# LLM-LEx runs}  \\ 
\hline
$\sqrt{| \sin (x)| }$ & $0.786005$ & 
$0.752 |\sin(x)|+0.302$ & 2  \\ 
\hline
$e^{1.83169 -\frac{3.35509}{x}}$ & $\mathcal{P}(x^9)$ & 
$\begin{array}{c}
 -2.586\tanh(1.159 x) + 3.73 \log(1.096 + x) \\-0.046 x^{1.568} - 0.371 
 \end{array}$
 & 2
\\ 
\hline
$x^3$ & \checkmark & \checkmark & 1  \\ 
\hline
$\left(\sqrt{x}+1.44439\right) (\log (x)+\pi )$ &
$\begin{array}{c}
-0.253251 - 2.97827\times2.1889^{-9.73907 x} \\+ 7.92679 \sqrt x + 33.2638 \times 3.24104^{-14.192 x}x
 \end{array}$
&
$\begin{array}{c}
 0.757 \sqrt{ x + 2.010} + 4.027 x^{0.631} \\+ 1.252 \log(x-0.0017) + 2.341
 \end{array}$
& 2
\\ 
\hline
$3.09529 x^3$ & \checkmark  & \checkmark & 1  \\ 
\hline
$\left(x^3+\pi \right)^2$ & \checkmark &

\checkmark

& 2
\\ 
\hline
$51.2288 \cos (1.18219 x)$ & $\mathcal{P}(x^{12})$ & \checkmark
& 1 \\ 
\hline
$-55.0512 \left(\sqrt{x}+1.\right)$ & \checkmark & \checkmark & 1\\ 
\hline
$x$ & \checkmark & \checkmark & 1 \\ 
\hline
$e^{\cos (x)}-0.0126997$ & $\mathcal{P}(x^{11})$ & 
$\begin{array}{c}
-0.115 \sin(2.221 x) + 0.307 \cos(1.809 x)\\ + 0.486 x^2 - 3.307 x+ 5.885\\ - 3.505 \exp(-1.075 x)
 \end{array}$
& 2 \\ 
\hline
$1.54251 -x$ & \checkmark & \checkmark & 1\\ 
\hline
$e^{2 x}$ & \checkmark & \checkmark & 1\\ 
\hline
$4.01209 +e^x$ & $\mathcal{P}(x^{12})$ & \checkmark & 1\\ 
\hline
$0.729202 \sqrt{x}-\pi$ & $\mathcal{P}(x^{10})$ & \checkmark & 1\\ 
\hline
$-3 x^3+x+1.99594$ & \checkmark & \checkmark & 1\\ 
\hline
$\log (x+1)$ & $\mathcal{P}(x^{7})$ & \checkmark & 1\\ 
\hline
$\sin \left(\log \left(\frac{4.1746}{x}\right)\right)$ &
$\begin{array}{c}
126.517 x^{4.00974}-178.8 x^{3.68092}\\
-0.00171611 x^{9.}+0.0512601 x^{8.}-0.583991 x^{7.}\\
+3.1446 x^{6.}-11.7513 x^{5.}+527.55 x\\
+0.232186 \log (x)-516.318 \sin (x)\\
+64.7972 \cos (x)-65.6805
\end{array}$
&
$\begin{array}{c}
1.299 e^{-0.182 x}\sin(0.391 x + 1.518)\\ -2.812 e^{-4.401 x}
 \end{array}$
& 2
\\ 
\hline
$\cos \left(e^x\right)+4.67315$ & $4.54681$ &

\checkmark
& 1
\\ 
\hline
$2 e^{-3 x}+e^{-x}$ & $\mathcal{P}(x^{11})$ &
$
\frac{5.796}{(x+4.673)^{9.427}} - 4.611 e^{-1.809 x} - 7.187\times 10^{-4}
$
& 2
\\ 
\hline
$\frac{x+4.11509}{x^3}$ & $4.22225/x^3$ &
$
\frac{0.778}{(x+0.009)^{2.143}} - 0.489 e^{-0.282 x} + \frac{0.422}{x-0.056}$
& 2
\\ 
\hline
\end{tabular}
\end{adjustbox}
\caption{$\mathcal{P}(x^n)$ indicates a polynomial of degree $n$, where the target function was not a polynomial, and \checkmark indicates the exact expression (to within numerical error of the coefficients). All runs with LLM-LEx had a population size of $25$ and ran for $10$ generations. The initial run had a terminal threshold of $10^{-7}$ with elitism, while the second had $10^{-10}$ without elitism. These runs can be found in the Jupyter notebook \texttt{fit\_functions.ipynb} in the examples folder of the GitHub repository~\cite{llmsr}. Little improvement is achieved with repeated iterations with Mathematica. In all cases the data were given by $100$ evenly spaced points with $x$ between $0.1$ and $5$. The scores from both methods are given in Table~\ref{tab:CompareScores}. Plots of four examples can be found in Figure~\ref{fig:examples}.}
\label{tab:Compare}
\end{table}
\end{landscape}

\begin{table}[h]
    \centering
\begin{tabular}{|c|c|c|}
\hline
{\bf Expression} & {\bf Mathematica Score}  & {\bf LLM-LEx Score} \\ 
\hline
$\sqrt{| \sin (x)| }$ & $-0.87$* & 
$-0.091$   \\ 
\hline
$e^{1.83169 -\frac{3.35509}{x}}$ & $-0.36$* & $-10^{-7}$
\\ 
\hline
$x^3$ & 0 & 0  \\ 
\hline
$\left(\sqrt{x}+1.44439\right) (\log (x)+\pi )$ &
$-10^{-6}$
&
$-10^{-11}$
\\ 
\hline
$3.09529 x^3$ & 0  & 0  \\ 
\hline
$\left(x^3+\pi \right)^2$ & 0 &
$0$
\\ 
\hline
$51.2288 \cos (1.18219 x)$ & $-10^{-13}$* & 0\\ 
\hline
$-55.0512 \left(\sqrt{x}+1.\right)$ & 0 & 0\\ 
\hline
$x$ & 0 & 0 \\ 
\hline
$e^{\cos (x)}-0.0126997$ & $-10^{-8}$* & 
$-10^{-5}$\\ 
\hline
$1.54251 -x$ & 0 & 0\\ 
\hline
$e^{2 x}$ & 0 & 0\\ 
\hline
$4.01209 +e^x$ & $-10^{-16}$* & 0\\ 
\hline
$0.729202 \sqrt{x}-\pi$ & $-10^{-8}$* & 0 \\ 
\hline
$-3 x^3+x+1.99594$ & 0 & 0\\ 
\hline
$\log (x+1)$ & $-10^{-7}$* & 0\\ 
\hline
$\sin \left(\log \left(\frac{4.1746}{x}\right)\right)$ &
$-10^{-5}$
&
$-10^{-4}$
\\ 
\hline
$\cos \left(e^x\right)+4.67315$ & $-0.02$* &
0
\\ 
\hline
$2 e^{-3 x}+e^{-x}$ & $-10^{-7}$* &
$-10^{-7}$
\\ 
\hline
$\frac{x+4.11509}{x^3}$ & $-0.1$ &
$-10^{-4}$
\\ 
\hline
\end{tabular}
\caption{Score comparison (less negative is better) for the functions in Table~\ref{tab:Compare}. Scores with a star indicate the cases where the traditional method returned a constant or high order polynomial (while the target expression was not), rather than a more interpretable representation. Plots of four examples can be found in Figure~\ref{fig:examples}.}
\label{tab:CompareScores}
\end{table}

\subsection{Using open models}

In this section, we aim to demonstrate that LLM-LEx can be used locally, even with modest resources, by employing open-source language models.

Specifically, we benchmark several open-source models against \texttt{gpt-4o}, by fitting the 20 functions listed in Table~\ref{tab:Compare}. All models are executed locally using \texttt{ollama 0.4.7}, with the following model configurations:
\begin{center}
\begin{tabular}{c|c|c}
     Model Name & \# Params & Size \\ \hline 
     \texttt{llama3.3} & 70B & 43 GB \\ 
     \texttt{mistral} & 7B & 4.1 GB \\ 
     \texttt{codestral} & 22B & 13 GB 
\end{tabular}
\end{center}
More specifically, we use \texttt{llama3.3:70b-instruct-q4\_K\_M}, the original \texttt{codestral}, and \texttt{mistral} v0.3. 
%Running each model requires several gigabytes beyond its stated size. 
All experiments were conducted on a 2024 M4 Max MacBook Pro, using a population size of 25 and evolving for 10 generations. The complete suite of 20 functions required approximately 4 hours (\texttt{codestral}, \texttt{mistral}), 7 hours (\texttt{llama3.3}), and 22 minutes (\texttt{gpt-4o}). In each LLM-LEx example, a single run was performed in asynchronous mode. The average scores and runtimes across all functions were as follows:
\begin{center}
\begin{tabular}{c|c@{\hskip 2em}c}
\textbf{Model} & \textbf{Avg Scores} & \textbf{Avg Time (s)} \\
\hline
\texttt{gpt-4o} & $-5.81 \times 10^{-3} \pm 2.51 \times 10^{-2}$ & $29.27 \pm 65.42$ \\
\texttt{llama3.3} & $-2.05 \times 10^{-2} \pm 8.95 \times 10^{-2}$ & $536.25 \pm 2922.17$ \\
\texttt{codestral} & $-2.63 \times 10^{-2} \pm 1.13 \times 10^{-1}$ & $202.46 \pm 768.27$ \\
\texttt{mistral} & $-2.79 \times 10^{-2} \pm 1.20 \times 10^{-1}$ & $187.10 \pm 702.53$ \\
\end{tabular}
\end{center}
As expected, \texttt{gpt-4o} is the fastest, benefiting from execution on OpenAI hardware, while \texttt{llama} (the largest local model) is the slowest; \texttt{codestral} and \texttt{mistral} exhibit comparable performance. However, the data is influenced by difficult outliers that introduce high variance, and all median scores are approximately $-10^{-17}$.

\begin{figure}[h]
    \centering
    \includegraphics[width=.7\textwidth]{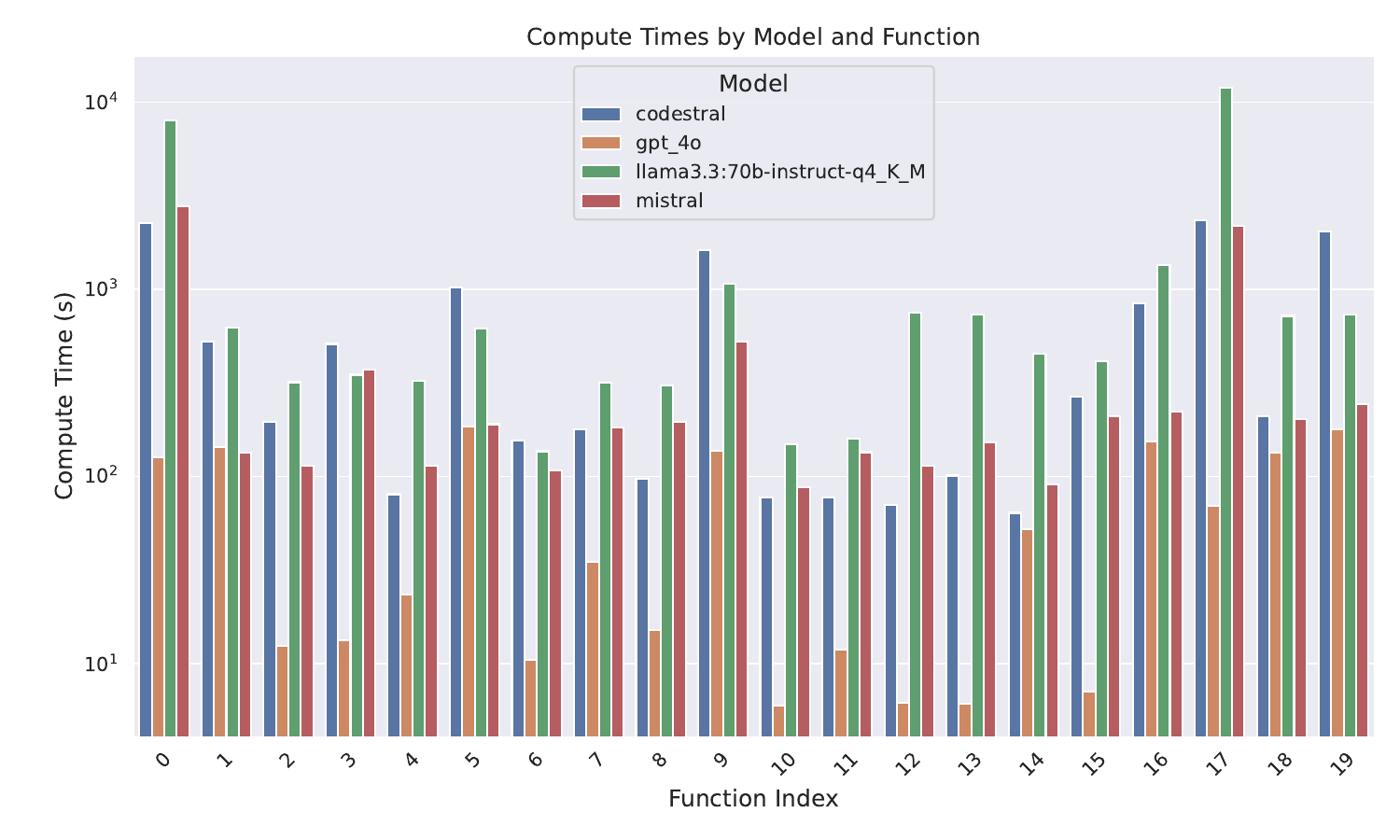} \\ 
    \includegraphics[width=.7\textwidth]{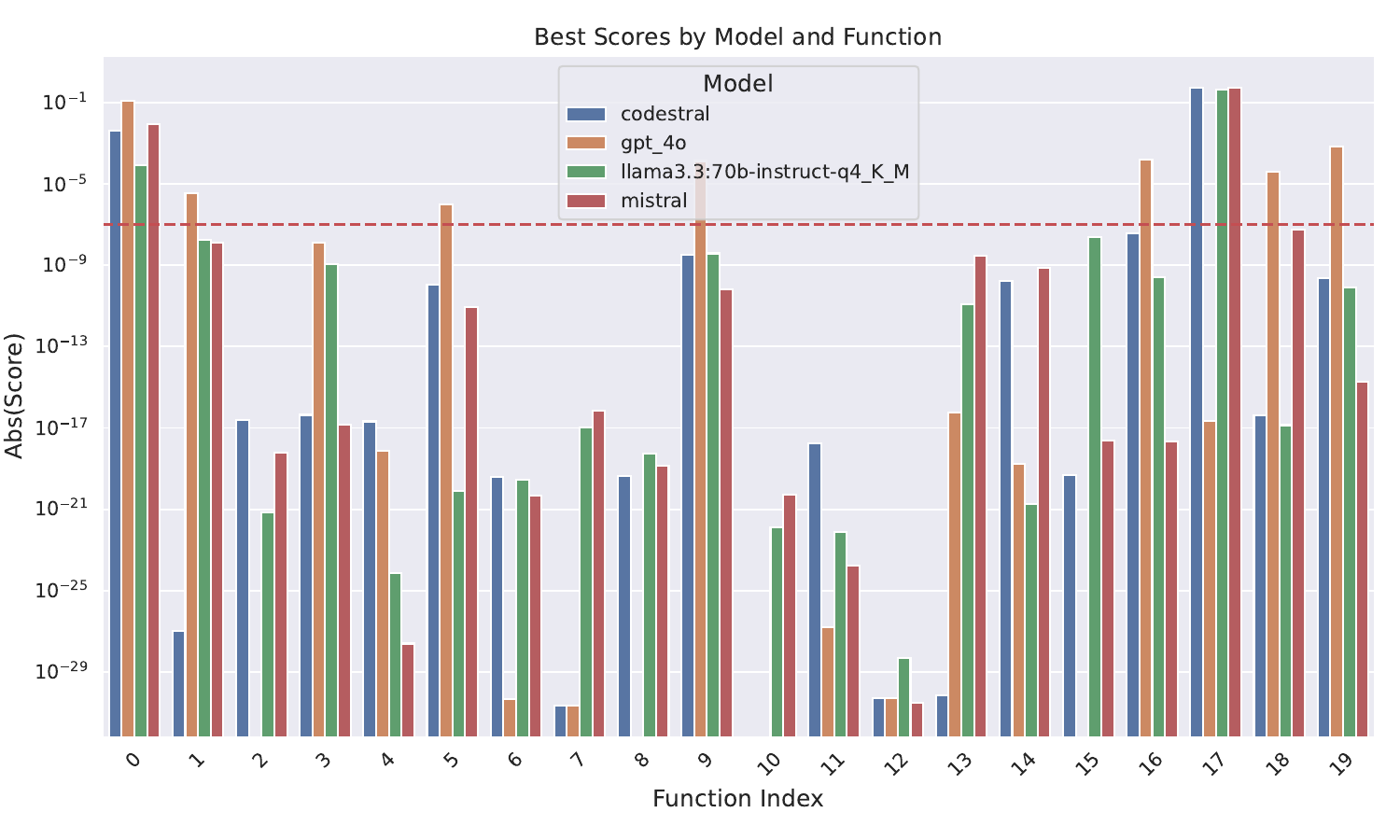} 
    \caption{Compute times and scores of LLM-LEx across different LLMs. Below the dashed red line, the genetic algorithm is stopped early due to high performance.}
    \label{fig:open_llms_times_and_scores}
\end{figure}

Function-by-function computation times and scores are shown in Figure~\ref{fig:open_llms_times_and_scores}. These plots highlight the substantial speed advantage of \texttt{gpt-4o}; nonetheless, in many cases, the open-source models are sufficiently fast — for example, \texttt{codestral} completes many tasks in under 100 seconds.

Table~\ref{tab:LLMCompare} presents the results indicating whether each model produced the correct expression. For this table, we first required a score greater than $-10^{-15}$ for an expression to be considered for correctness evaluation. The remaining expressions were then manually inspected, allowing the application of trigonometric identities in verifying correctness. From the table, it is evident that \texttt{codestral} performs very well, achieving results comparable to \texttt{gpt-4o}.

\medskip
\begin{figure}[t]
    \centering
    \includegraphics[width=.9\textwidth]{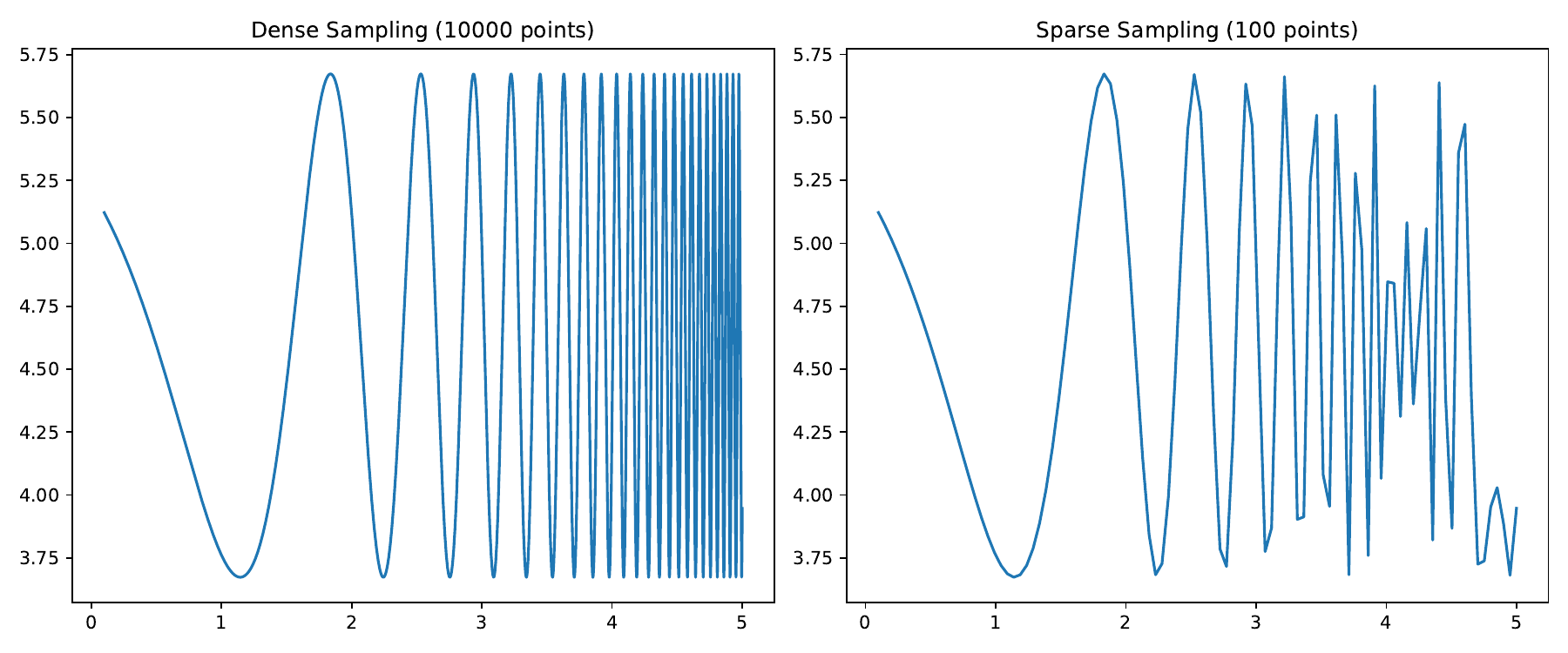}
    \caption{Dense vs. Sparse sampling of $x$ values for $f(x) = 4.67315 + \cos(\exp(x))$.}
    \label{fig:dense_vs_sparse_sampling_complicated}
\end{figure}
\emph{Higher resolution for harder examples.} In the previous analysis of open-source models, one of the more challenging cases was the function $f(x) = 4.67315 + \cos(\exp(x))$, where the corresponding image was generated using 100 evenly spaced $x$-values in the interval$(0.1,5)$. For larger $x$, the function oscillates rapidly, and the sparse sampling results in interpolation artefacts due to aliasing errors. The impact of this aliasing is illustrated in Figure~\ref{fig:dense_vs_sparse_sampling_complicated} and is clearly substantial. Given the severity of the aliasing, it is remarkable that LLM-LEx with \texttt{gpt-4o} was able to fit the function.

To assess the significance of the aliasing effect, we refitted LLM-LEx using a denser sampling of points, as shown on the left-hand side of the figure. In this case, \texttt{mistral} achieved a perfect fit, whereas the other models failed.
As \texttt{gpt-4o} executes considerably faster, we repeated the same experiment four additional times; it succeeded in three out of four runs, yielding a $4/5$ total success rate. Given the probabilistic nature of language models at non-zero temperature, this variability is unsurprising. 

\begin{table}[t]
\begin{adjustbox}{width=\columnwidth,center}
\begin{tabular}{|c|c|c|c|c|c|}
\hline
\textbf{Expression} & \textbf{Mathematica} & \textbf{gpt-4o} & \textbf{llama3.3} & \textbf{mistral} & \textbf{codestral} \\
\hline
$\sqrt{| \sin (x)| }$ &  & & & & \\
\hline
$e^{1.83169 -\frac{3.35509}{x}}$ &  & & & &\checkmark \\
\hline
$x^3$ & \checkmark &\checkmark &\checkmark & \checkmark& \checkmark\\
\hline
$\left(\sqrt{x}+1.44439\right) (\log (x)+\pi )$ &  & & & & \\
\hline
$3.09529 x^3$ & \checkmark &\checkmark &\checkmark & \checkmark& \checkmark\\
\hline
$\left(x^3+\pi \right)^2$ & \checkmark & & & & \\
\hline
$51.2288 \cos (1.18219 x)$ & &\checkmark &\checkmark & &\checkmark \\
\hline
$-55.0512 \left(\sqrt{x}+1.\right)$ & \checkmark & \checkmark& & \checkmark&\checkmark \\
\hline
$x$ & \checkmark & \checkmark&\checkmark &\checkmark &\checkmark \\
\hline
$e^{\cos (x)}-0.0126997$ &  & & & &\\
\hline
$1.54251 -x$ & \checkmark & \checkmark& \checkmark& \checkmark&\checkmark \\
\hline
$e^{2 x}$ & \checkmark &\checkmark & \checkmark&\checkmark & \checkmark\\
\hline
$4.01209 +e^x$ &  &\checkmark & &\checkmark & \checkmark\\
\hline
$0.729202 \sqrt{x}-\pi$ &  &\checkmark & & &\checkmark \\
\hline
$-3 x^3+x+1.99594$ & \checkmark & \checkmark& \checkmark& & \\
\hline
$\log (x+1)$ &  &\checkmark & &\checkmark &\checkmark \\
\hline
$\sin \left(\log \left(\frac{4.1746}{x}\right)\right)$ &  & & & & \\
\hline
$\cos \left(e^x\right)+4.67315$ &  &\checkmark & & & \\
\hline
$2 e^{-3 x}+e^{-x}$ &  & &\checkmark & &\checkmark \\
\hline
$\frac{x+4.11509}{x^3}$ &  & & & & \\
\hline
\textbf{Total Correct} & 8 & 12 & 8 & 8 & 12\\ \hline
\end{tabular}
\end{adjustbox}
\caption{Comparison of Mathematica results with four different LLMs utilised by LLM-LEx. The best results utilised LLM-LEx with \texttt{gpt-4o} or \texttt{codestral}. All runs with LLM-LEx were single shot.}
\label{tab:LLMCompare}
\end{table}

\subsection{Adding noise}\label{sec:noise}
Thus far, LLM-LEx has been tested exclusively on clean data, free from noise. However, such conditions are rarely encountered in scientific domains. We therefore briefly explore how LLM-LEx responds to the introduction of noise.

As examples, we consider four of the functions that LLM-LEx successfully identified in the absence of noise in the previous section, namely:
\begin{align*}
    x^3&, \quad 51.2288 \cos (1.18219 x),\quad x,\quad 4.01209 +e^x 
\end{align*}
We add noise $\xi$, where
\begin{align}
    y &= y_\text{true} + \xi,\\
    \xi &\sim \mathcal{N}(0,\epsilon \, |y_\text{max}|), 
\end{align}
and $\mathcal{N}(\mu,\sigma)$ is a normal distribution with mean $\mu$ and standard deviation $\sigma$.

When introducing noise into our curve fitting algorithm, we must naturally increase the acceptable error tolerance (the exit condition), as the model should not be expected to match the ground truth as closely as with noise-free data. Without this adjustment, we observed that LLM-LEx tends to identify functions that, while visually approximating the ground truth well, on closer inspection instead overfit the noise. By contrast, the single-generation (single-shot) ansatz, which involves fewer parameters, offers a reasonably accurate approximation of the true function. The success of the single-shot approach may indicate that the LLM, in a sense, recognises it is working with noisy data and prioritises fitting the underlying ground truth rather than the noise. Indeed, we can simply request that it attempt to do this. Nevertheless, as the noise level increases, the probability of recovering the exact ground truth function diminishes.

Let us now present two illustrative examples. In both cases, the \( x \)-values consist of 100 points uniformly spaced between 0.1 and 5.0. We considered noise levels \( \epsilon \in \{0.01, 0.03\} \) and set the exit condition to 0.04.

In the first example, we fit the function \(f(x) = 51.2288 \cos(1.18219x) \). LLM-LEx returned the correct expression for \( \epsilon = 0 \) and 0.01, but for \( \epsilon = 0.03 \), it fit the data to an exponential function, indicating that the noise had become too large for reliable recovery of the true form.

In the second example, we fit the simple linear function \( f(x) = x \). Again, LLM-LEx returned the correct result for \( \epsilon = 0 \) and 0.01. However, for \( \epsilon = 0.03 \), it produced the expression
\begin{equation}
2.17\,x - 11.2 \sin(0.109x) + 0.053.
\end{equation}
While this initially appears incorrect, a Taylor expansion of the low-frequency sine term yields
\begin{equation}
0.053 + 0.952x + 0.00243x^3 - 1.45 \times 10^{-6}x^5 + O(x^6).
\end{equation}
Examining these terms over the domain, it is evident that the dominant contribution beyond the constant is \( 0.952x \), with the higher-order terms negligible and on the order of the noise. In effect, despite the seemingly more complex ansatz, the model is still approximating a linear function. This suggests that LLM-LEx, in the presence of moderate noise, tends to prioritise capturing the underlying trend of the true function, even if the explicit form deviates from the exact ground truth.

\subsection{Special functions and prompt engineering}
One potential advantage of LLM-LEx over traditional symbolic regression is that it does not require the user to specify a predefined basis of functions. When provided with access to appropriate Python packages, LLM-LEx can even propose special functions within its suggested ansätze. 

As an example, we can modify the prompt (the change to the default prompt is underlined here for emphasis) to:

\begin{center}

Prompt 1:\begin{it} ``You are a symbolic regression expert. Analyze the data in the image and provide an improved mathematical ansatz.
 Respond with ONLY the ansatz formula, without any explanation or commentary. Ensure it is in valid Python. 
 You may use numpy functions\underline{, and scipy.special}. 
 params is a list of parameters that can be of any length or complexity.''
 \end{it}
\end{center}

Given this prompt, we found that LLM-LEx could readily identify certain special functions, such as the error function, with relative ease. However, for more complex combinations of special functions, LLM-LEx typically produced accurate approximations using functions from \texttt{numpy}, unless the prompt was adjusted to explicitly emphasise the use of \texttt{scipy.special}. Specifically, we consider the following prompt:

\begin{center}

Prompt 2: \begin{it}``You are a symbolic regression expert. Analyze the data in the image and provide an improved mathematical ansatz.
 Respond with ONLY the ansatz formula, without any explanation or commentary. Ensure it is in valid Python. 
 You may use numpy functions\underline{, and scipy.special. Give preference to scipy.special over numpy}. 
 params is a list of parameters that can be of any length or complexity.''
 \end{it}
\end{center}

In addition to varying the prompts, we also modified the \( x \)-values used to generate the image, aiming to push beyond the domains where the \texttt{numpy} approximate expressions are valid. Specifically, we define \texttt{xVals1} to be 100 uniformly spaced points between -3 and 3, while \texttt{xVals2} as 500 uniformly spaced points between -10 and 10.

In the second example, we study the Bessel function \( J_0 \). For this case, Prompt 1 yields a score of order \( 10^{-9} \) for \texttt{xVals1} and \( 10^{-4} \) for \texttt{xVals2}, with the plot of the LLM-LEx function and the ground truth being indistinguishable to the naked eye. The returned expressions, however, are approximations of the Bessel function on this interval, consisting of \texttt{numpy} trigonometric functions and polynomials. A similar outcome is observed for Prompt 2 with \texttt{xVals1}. However, for Prompt 2 and \texttt{xVals2}, the score improves to \( 10^{-15} \), with a perfect visual match, and LLM-LEx suggests \texttt{scipy.special.jn(0, x)}, which is precisely the desired Bessel function.

We will briefly explore further possibilities for prompt engineering when discussing extensions to this work in Section~\ref{sec:conc}.

\section{Combining with Kolmogorov-Arnold Networks}\label{sec:KANs}
\subsection{Univariate is all you need}\label{sec:UnivariateIAUN}
\begin{figure}[H]
\centering
\resizebox{0.5\textwidth}{!}{%
\begin{circuitikz}
\tikzstyle{every node}=[font=\LARGE]

% Fit KAN Box (Rounded)
\draw[rounded corners=8pt] (3.75,10.25) rectangle node {\LARGE Fit KAN} (7.5,7.75);

% Decision Diamond (Rounded manually)
\draw[rounded corners=8pt]
  (10,9) -- (13.75,7) -- (17.5,9)
  -- (13.75,10.75) -- cycle;
\node [font=\LARGE] at (13.75,9) {Prune Suggested?};

% Loop from decision "Yes" to Fit KAN (fully rounded)
\draw[rounded corners=12pt]
  (13.75,10.75) -- (13.75,12.75)
  -- (5.75,12.75)  -- (5.75,10.25);;

% Labeled arrow
\node [font=\LARGE] at (9.75,13.25) {Yes - update architecture };

% Fit KAN input arrow
\draw[->, >=Stealth] (1.25,9) -- (3.75,9);
\node [font=\LARGE] at (0.25,9) {Data};

% Fit KAN to decision
\draw[->, >=Stealth] (7.5,9) -- (10.15,9);

% Decision "No" path
\draw[->, >=Stealth] (13.75,7.1) -- (13.75,2.75);
\node [font=\LARGE] at (14.5,5.25) {No};

% Fit edges with LLM_LEx box (Rounded)
\draw[rounded corners=8pt] (9.25,2.75) rectangle node {\LARGE Fit edges with LLM-LEx} (17.75,0.25);

% Down to Simplify Expression
\draw[->, >=Stealth] (13.75,0.25) -- (13.75,-2.25);

% Simplify Expression box (Rounded)
\draw[rounded corners=8pt] (10,-2.25) rectangle node {\LARGE Simplify Expression} (17.5,-4.75);

\end{circuitikz}
}%
\caption{The general structure of KAN-LEx.}
\label{fig:kanlex}
\end{figure}
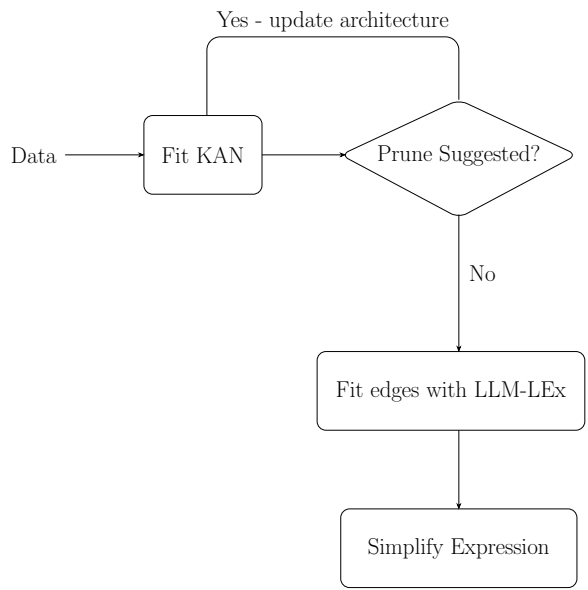
The visual symbolic regression technique we have introduced is developed for univariate functions $f:\mathbb{R}\to\mathbb{R}$ and utilises 2D images. However, it is not applicable to multivariate functions, since these cannot be plotted in 2D. 

Luckily, all we ever need from a mathematical point of view are (sums of) univariate functions. This is due to the Kolmogorov-Arnold representation theorem, which postulates that (under mild assumptions) any function can be written as sums of univariate functions\footnote{One should keep in mind, however, that there is no theoretical guarantee the required univariate functions are simple or remotely well behaved. In practice however, this does not appear to be an obstacle---particularly when we generalise to `deep' KANs.}. This fact was used in Kolmogorov-Arnold Networks (KANs)~\cite{Liu:2024swq}. The general structure of our approach is indicated in Fig.~\ref{fig:kanlex}, and is called KAN-LEx.

A KAN is a directed computational graph with edges between nodes that are arranged in layers, much like a standard MLP. Unlike an MLP, however, the edges of the graph that connect nodes $i$ of layer $\ell$ to node $j$ of layer $(\ell+1)$ are instead univariate functions of the value of node $i$ in layer $\ell$. In KANs, these functions are approximated by (cubic) splines. The operation on nodes is just the sum over the values from the incoming edges. Deep KANs are obtained by stacking multiple KAN layers, i.e., by interweaving sum operations and applying univariate functions to the sums. Thus, for a KAN with $L$ layers where each layer has only a single node, the sum at the nodes is just the identity operation, and the KAN represents a concatenation of univariate functions.

We can thus combine symbolic regression on univariate functions with KANs by applying the technique outlined in Section~\ref{sec:LLMLEX} to the univariate function on each edge of the KAN. In more detail, we proceed as follows.

Firstly, we define a KAN architecture of maximal desired complexity, meaning whose number of layers equals or exceeds the number of nested functions we expect or want to map to, and whose number of nodes exceeds the number of summands expected in the result. We should err on the over-parametrised side, since this can be more easily remedied. We can detect if we have too many layers if some of the edge functions are just linear (or affine) functions; we can then reduce the number of layers. The number of nodes per layer can be pruned efficiently using KANs built-in pruning routines which rely on edge scores, see~\cite{Liu:2024swq} for details.

Once we have settled on an architecture, we fit each univariate edge function as in Section~\ref{sec:LLMLEX}. Armed with an expression for each edge, we can build the nested function expressed by the KAN by constructing the expression from the computational graph. We simplify the expression with sympy~\cite{sympy} and refit the coefficients using scipy's optimisers. 

Next, we run the simplified expression through an LLM, asking it to further simplify using the following strategies:
\begin{itemize}
    \item Taylor expand terms that are small in this interval
    \item Remove negligible terms
    \item Recognise polynomial patterns as Taylor series terms
    \item Combine similar terms and factor them when possible
\end{itemize}
In order for the LLM to be able to judge which terms are small, negligible, or could be involved in an (inverse) Taylor expansion, we also provide the interval from which the input values were sampled. We then refit the free-parameters of the simplified expression.

The reason for this simplification step (\texttt{sympy} $\rightarrow$ \texttt{scipy} $\rightarrow$ \texttt{llm} $\rightarrow$ \texttt{scipy}) is that numerical fitting and function composition do not necessarily commute, and we can improve the quality of the fit by fitting the combination. At the very least, the constant terms $b_i^{\ell}$ appearing in the fit of each individual edge function is arbitrary, only their sum $\sum_i b_i^{\ell}$ is meaningful. Moreover, since KANs have to express everything in terms of sums of functions, some elementary operations are expressed in a cumbersome way by the KAN. For example, since KANs have no multiplication built in, they would express (for positive $x,y$)
\begin{align}
\label{eq:MultiplicationExample}
xy=\exp[\log(x)+\log(y)]\,,
\end{align}
meaning they would learn a log on the edges of the first layer, then sum the logs on the node, and then learn an exponential at the second layer. The simplification step (sympy, scipy, llm, scipy) can be repeated $N$ times, but we only used $N=1$ in our examples.

\subsection{Multivariate Examples}\label{sec:KANmulti}

We now demonstrate the efficacy of KAN-LEx in a number of examples.

\subsubsection*{Example 1}
\begin{figure}[t]
    \centering
    \subfloat[{$[2,4,4,1]$ (pruned) \label{fig:Ex1KAN2441}}]{\includegraphics[width=0.25\textwidth]{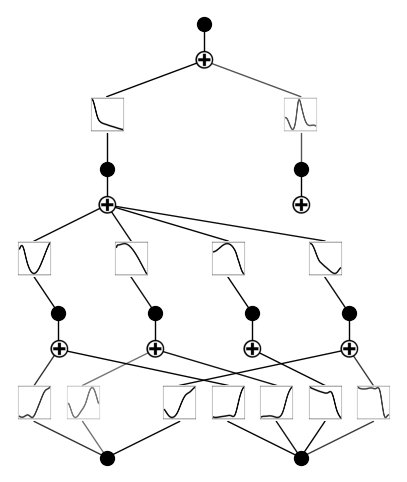}}
    \subfloat[{$[2,4,1,1]$ (pruned) \label{fig:Ex1KAN2411}}]{\includegraphics[width=0.25\textwidth]{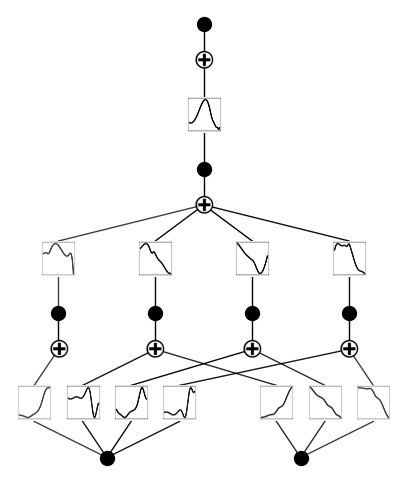}}
    \subfloat[{$[2,4,1]$ (pruned) \label{fig:Ex1KAN241}}]{\includegraphics[width=0.25\textwidth]{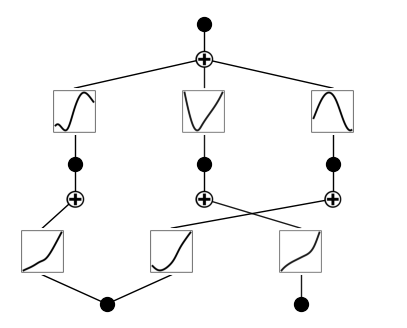}}
    \subfloat[{$[2,3,1]$ (pruned) \label{fig:Ex1KAN231}}]{\includegraphics[width=0.25\textwidth]{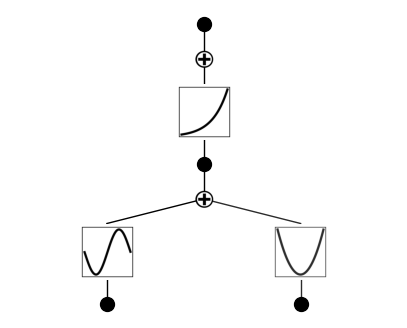}}
    \caption{Selection stages for the KAN architecture. The final model is a $[2,1,1]$ KAN.}
    \label{fig:Example1KANs}
\end{figure}

To illustrate the procedure of fitting multivariate functions, we choose the target
\begin{align}
\begin{array}{crl}
    f:\quad  &\mathbb{R}^2&\to\mathbb{R}\\
        & (x,y)&\mapsto \exp[\sin(\pi x) + y^2]
\end{array}
\end{align}
The first (and arguably most important) step is to find a good KAN architecture. We generate 10k points drawn random uniformly from the interval $-1\leq x,y\leq 1$. As a first guess for the KAN architecture, we choose a [2,4,4,1] KAN and prune it. The pruned model suggests an architecture choice of [2,4,1,1], see Figure~\ref{fig:Ex1KAN2441}. We retrain a [2,4,1,1] KAN and prune it. Pruning did not simplify the KAN but it has several edges with almost linear activation functions (see Figure~\ref{fig:Ex1KAN2411}), which motivates stripping a layer and retraining a [2,4,1] KAN. After pruning, we see that the architecture can be further simplified to a [2,3,1] KAN, cf.\ Figure~\ref{fig:Ex1KAN241}. Retraining and pruning a [2,3,1] KAN then gives the [2,1,1] KAN in Figure~\ref{fig:Ex1KAN231}.

In the next step, we fit a function to each edge. We use a population of 3 and 2 generations. The best-fitting LLM suggestions for the three edges $e^{(l)}_{ij}$ that connect node $i$ in layer $\ell$ to node $j$ in layer $(\ell+1)$ are
\begin{align}
\begin{array}{lll}
    e^{(1)}_{11}(x,y) &= a_0  \sin(a_1 x) + a_2 \cos(a_3 x)\,,\quad&\vec{a}\approx (0.373, 3.142, 0.170,  10^{-6})\\
    e^{(1)}_{21}(x,y) &= b_0  y^2 + b_1 y + b_2\,,\quad&\vec{b}\approx (0.373,  10^{-5}, -0.166)\\
    e^{(2)}_{11}(z) &= c_0 + c_1 z + c_2 * \exp(c_3 z)  \,,\quad&\vec{c}\approx (-0.031, -0.050, 1.018, 2.651) 
\end{array}
\end{align}

Although the functions are close to the correct expressions, they show some of the subtleties discussed in Section~\ref{sec:UnivariateIAUN}. Firstly, the argument of the cosine is (almost) zero, meaning that the edge function is $e^{(1)}_{11} \approx a_0  \sin(a_1 x) + a_2$. The constant $a_2\approx-b_2$, meaning that the two constants will cancel out upon summing the contributions at the node.  Second, the coefficient $a_0$ in front of the sine term is equal to the coefficient $b_0$ in front of the quadratic term, and both are roughly equal to $1/c_3\approx a_0\approx b_0$. The linear term in the second edge function is close to zero, and the linear term in the last edge function
\begin{align}
    c_1 z \approx c_1 (a_0  \sin(a_1 x) + b_0  y^2)
\end{align}
is small with $c_1 a_0\approx c_1 b_0\approx 10^{-2}$.

This discussion illustrates the necessity of rounding coefficients to get rid of terms that are close to zero, simplifying the expression with sympy and an LLM, and refitting the parameters with scipy. This pipeline automatises the discussion in the previous paragraph. The sympy and LLM optimiser approximate the cosine term with just $a_2$ and combine it with $b_2$. Moreover, they combine $c_1 a_0$ and $c_1 b_0$. Refitting the resulting expression with scipy and neglecting terms that are numerically zero, we get the function (rounded to six significant digits)
\begin{align} 
    1.0 \exp[1.0 y^2 + \sin(3.141593x)]\,.
\end{align}

\subsubsection*{Example 2}
As a second example, we use the multiplication example in~\eqref{eq:MultiplicationExample} and study the range $1\leq x,y\leq2$. We start again with a [2,4,4,1] KAN, which after pruning suggests a [2,1,1] architecture. Running edge fitting on this architecture gives
\begin{align}
\begin{array}{lll}
    e^{(1)}_{11}(x,y) &= a_0  \log(x) + a_1\,,\quad&\vec{a}\approx (1.182, 0.261)\\
    e^{(1)}_{21}(x,y) &= b_0  \log(y) + b_1\,,\quad&\vec{b}\approx (1.182, 0.339)\\
    e^{(2)}_{11}(z) &=  c_0 \exp(c_1 z) \,,\quad&\vec{c}\approx (0.602, 0.846) 
\end{array}
\end{align}
We find again that $a_0\approx b_0\approx 1/c_0$. There is also a discrepancy in the constant terms, $a_1\neq -b_1$, but also $c_0\neq 1$. In fact, $c_0\approx 1/e^{c_1(a_1+b_1)}$, such that the final functional form is
\begin{align}
\begin{split}
    f(x,y)&=c_0 e^{c_1 (a_0\log(x)+b_0\log(y)) + c_1(a_0+b_0)}\\
    &\approx c_0 e^{c_1(a_0+b_0)} e^{\log(x)+\log(y)}\\
    &\approx e^{\log(x)+\log(y)}\,.
\end{split}
\end{align}

\begin{figure}[t]
    \centering
    \subfloat[{$[2,4,4,1]$ (pruned) \label{fig:Ex2KAN2441}}]{\includegraphics[width=0.4\textwidth]{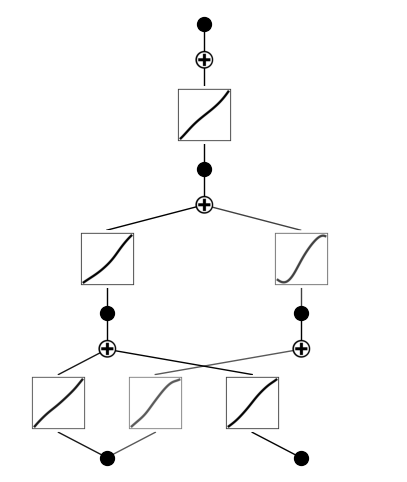}}
    \subfloat[{$[2,1,1]$ (pruned) \label{fig:Ex2KAN2411}}]{\includegraphics[width=0.4\textwidth]{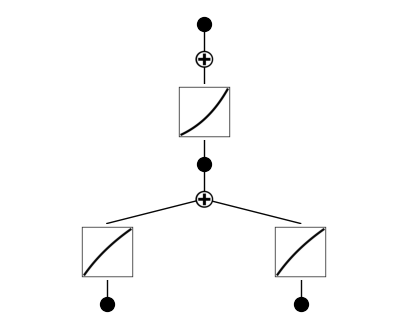}}
    \caption{Selection stages for the KAN architecture. The final model is a $[2,1,1]$ KAN.}
    \label{fig:Example2KANs}
\end{figure}

Of course these simplifications are discovered automatically in our pipeline. The scipy refitting does not change much, since the KAN is already very accurate. The sympy simplification step turns $e^{\log(x)+\log(y)}$ into $xy$, and the subsequent fitting step does not change anything. It is interesting to see what the subsequent LLM call suggests given that it is already presented with a solution. We ask for 9 proposals to improve the function, and it returns
\begin{align}
\begin{split}
    &\texttt{[`x~*~y',~`y~*~x', `x~*~y~+~0', `x*y', `y*x'}\\ 
    &\texttt{'x**1~*~y**1','x*y', `x**1~*~y**1', `x*y~+~0']}\,,
\end{split}
\end{align}
so it returns 9 equivalent expressions (out of which only 7 are distinct strings). In any case, we now have the correct function, and the final scipy fitting step does of course not change anything.
\subsubsection*{Example 3 - Newton's Law}
As a third, and final, example we consider Newton's law of gravitation, given by \( V(r) = \frac{GMm}{r} \). In dimensionless form, this reduces to the expression \( xy/z \). We examine the range of variables \( 0.5 < x, y, z < 3 \), and begin by training a KAN with architecture \([3, 4, 4, 1]\). Several rounds of pruning yield the sequence:
\[
[3,4,4,1] \rightarrow [3,2,2,1] \rightarrow [3,1,1] \rightarrow [3,1],
\]
at which point we fit each edge using LLM-LEx, and find
\begin{align}
\begin{array}{lll}
    e^{(1)}_{11}(x) &= a_0 \log(x) + a_1\,,\quad&\vec{a}\approx (2.026, -0.605)\\
    e^{(1)}_{21}(y) &= b_0  \log(y) + b_1\,,\quad&\vec{b}\approx (2.029, -6.302)\\
    e^{(1)}_{31}(z) &= c_0  \log(z) + c_1\,,\quad&\vec{c}\approx (-2.029, -1.547)\\
    e^{(2)}_{11}(w) &=  d_0 \exp(d_1 w) + d_2 \,,\quad&\vec{d}\approx (64.608, 0.493 ,-0.001).
\end{array}
\end{align}
Simplifying and refitting automatically returns the correct function $x y / z$.

\subsection{Improving Univariate Examples}
The methodology outlined above can also be applied to enhance our univariate results. Starting with data for a univariate function, we follow the approach described in the previous subsection: pruning the architecture, fitting a function to each edge of the pruned KAN, and subsequently simplifying the result.

Although this approach is clearly feasible and intuitively appealing---particularly for ``deep'' functions (i.e., functions that are naturally expressed as compositions of many simpler functions)—we found it to be largely ineffective in practice. We applied it to those functions in Table~\ref{tab:Compare} for which LLM-LEx failed to identify the exact expression, but only one case, $2 e^{-3 x}+e^{-x}$, yielded a successful outcome.

\section{Conclusion}\label{sec:conc}
In conclusion, we have presented a new method for univariate symbolic regression based on large language models and \texttt{funsearch}. This approach has demonstrated remarkable success across a wide range of functions, despite the simplicity of the underlying genetic algorithm. Notably, we do not enforce a simplicity score, which is often employed to reduce bloat, suggesting that the LLM inherently exhibits a bias towards simplicity. The results are summarised in Tables~\ref{tab:Compare} and~\ref{tab:CompareScores}. Furthermore, as discussed in Section~\ref{sec:noise}, we observed that the method exhibits a degree of resilience to moderate noise, albeit with a tendency to overfit if the noise is too large.

In Section~\ref{sec:KANs}, we extend our univariate method to multivariate functions by first training a KAN on the dataset. This process decomposes the multivariate problem into several univariate subproblems. We then apply our previous method to each of these subproblems individually, before simplifying the combined expression (using the assistance of LLMs as well as standard symbolic simplification techniques as implemented in \texttt{sympy}).

We have developed a Python package that facilitates applications to both univariate and multivariate cases, which is available on GitHub~\cite{llmsr}. The GitHub repository also includes the Jupyter Notebooks containing the benchmarks and examples discussed in this paper.

There are several natural extensions to this work. One obvious direction is to explore more sophisticated genetic algorithms as an enhancement to LLM-LEx. Another promising avenue involves training purpose-built vision transformers on images of known functions. One might consider training higher-dimensional analogues of vision transformers to handle multivariate functions directly, thereby bypassing the need for KANs. However, the latter two approaches would require constraining the symbolic regression task to a specific class of functions. In contrast, a commercial LLM is likely to have encountered a far broader range of functional forms during pretraining; it follows that an effective route might be to fine-tune models on symbolic regression data. In domain-specific contexts, where the types of functions expected are more predictable, this limitation could potentially be mitigated by training the transformer exclusively on relevant function classes. 

Finally, another avenue worth exploring is prompt engineering---in particular, tailoring the LLM's prompt to incorporate domain-specific knowledge. For instance, if the target function represents a particular physical quantity, the LLM might be able to leverage insights from scientific domains in suggesting appropriate functions.

\section*{Acknowledgements}
KFT is supported by the Gould-Watson Scholarship. JH, TRH, and FR are supported by the National Science Foundation under Cooperative Agreement PHY-2019786 (The NSF AI Institute for Artificial Intelligence and Fundamental Interactions, http://iaifi.org/). FR is also supported by the NSF grant PHY-2210333 and startup funding from Northeastern University. JH is supported by NSF CAREER grant PHY-1848089.

\bibliographystyle{ytphys}
\bibliography{sn-bibliography}

\end{document}